\def\year{2020}\relax
\documentclass[letterpaper]{article} % DO NOT CHANGE THIS
\usepackage{cite}
\usepackage{aaai20}  % DO NOT CHANGE THIS
\usepackage{times}  % DO NOT CHANGE THIS
\usepackage{helvet} % DO NOT CHANGE THIS
\usepackage{courier}  % DO NOT CHANGE THIS
\usepackage[hyphens]{url}  % DO NOT CHANGE THIS
\usepackage{graphicx} % DO NOT CHANGE THIS
\usepackage{graphicx}  % DO NOT CHANGE THIS
\usepackage{multirow}
\usepackage{color}
\usepackage{makecell}
\title{Logo-2K+: A Large-Scale Logo Dataset for Scalable Logo Classification}
\author{\textbf{Jing Wang$^{1}$, Weiqing Min$^{2}$, Sujuan Hou$^{1}$, Shengnan Ma$^{1}$, Yuanjie Zheng$^{1}$, Haishuai Wang$^{3}$, Shuqiang Jiang$^{2}$}\\
\textsuperscript{\rm 1} School of Information Science and Engineering, Shandong Normal University\\
\textsuperscript{\rm 2} Institute of Computing Technology, Chinese Academy of Sciences \\
\textsuperscript{\rm 3} Department of Computer Science and Engineering, Fairfield University\\}

\begin{document}
\maketitle
\begin{abstract}
Logo classification has gained increasing attention for its various applications, such as copyright infringement detection, product recommendation and contextual advertising. Compared with other types of object images, the  real-world logo images have larger variety in logo appearance and more complexity in their background. Therefore, recognizing the logo from images is challenging. To support efforts towards scalable logo classification task, we have curated a dataset, Logo-2K+, a new large-scale publicly available real-world logo dataset with 2,341 categories and 167,140 images. Compared with existing popular logo datasets, such as FlickrLogos-32 and LOGO-Net, Logo-2K+ has more comprehensive coverage of logo categories and larger quantity of logo images. Moreover, we propose a Discriminative Region Navigation and Augmentation Network (DRNA-Net), which is capable of  discovering more informative logo regions and augmenting these image regions for logo classification. DRNA-Net consists of four sub-networks: the navigator sub-network first selected informative logo-relevant regions guided by the teacher sub-network, which can evaluate its confidence belonging to the ground-truth logo class. The data augmentation sub-network then augments the selected regions via both region cropping and region dropping. Finally, the scrutinizer sub-network fuses  features from augmented regions and the whole image for logo classification. Comprehensive experiments on Logo-2K+ and other three existing benchmark datasets demonstrate the effectiveness of proposed method. Logo-2K+ and the proposed strong baseline DRNA-Net are expected to further the development of scalable logo image recognition, and the Logo-2K+ dataset can be found at {\color{red}{\url{https://github.com/msn199959/Logo-2k-plus-Dataset}}}.
\end{abstract}

\section{Introduction}
Logo classification~\cite{Jingying-NLR-PR2003,Hoi2015LOGO,Bianco2017Deep,II2019Scalable} aims to recognize the logo name of an input image, and can be viewed as a special case of image recognition from a computer vision perspective. It has been a well-studied subject for decades for its various real-world commercial applications, such as brand retrieval~\cite{Joly2009Logo}, brand monitoring~\cite{Liu2018Visual,Liao2017Representativeness}, trademark infringement detection, product recommendation and intelligent transportation.
\begin{figure*}[!htbp]
\centering
\includegraphics[width=0.95\textwidth]{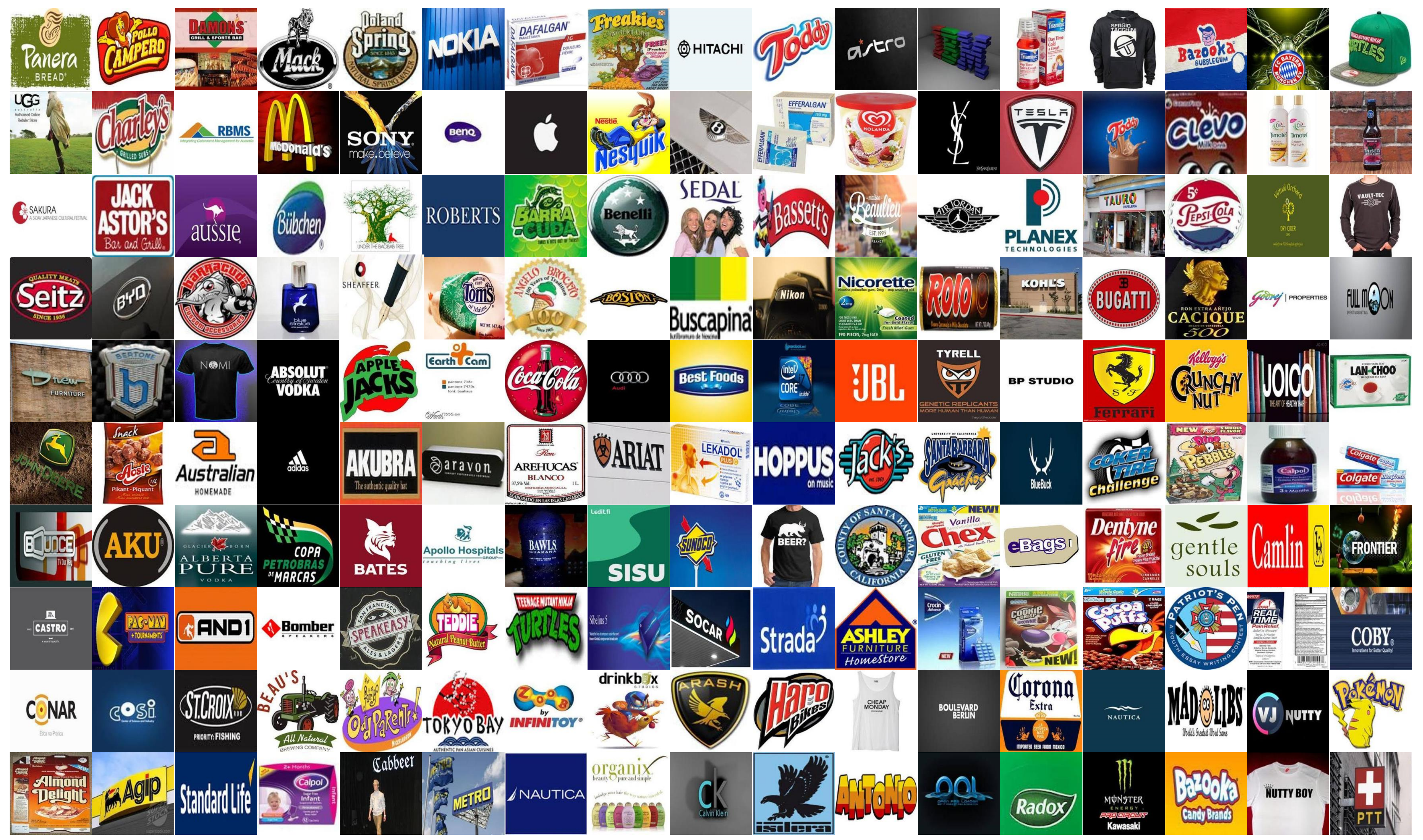}
\caption{Some logo images from Logo-2K+ dataset.}
\label{fig1}
\end{figure*}

Traditional methods for logo classification rely on hand-crafted features and keypoint-based detectors~\cite{Romberg2013Bundle}. Recently deep learning has been used for logo classification for its superior performance with end-to-end pipeline automation~\cite{Hoi2015LOGO,Bianco2017Deep,II2019Scalable}. One standard approach is that one image is fed into one deep neural object detector (e.g., fast R-CNN~\cite{Girshick2015Fast}) and one classifier outputs its prediction, where  the ImageNet trained CNN is fine-tuned on FlickrLogos-32~\cite{Romberg2011Scalable} dataset. However, such methods suffer from the lack of a big quality logo dataset, leading to lower generalization ability of these models. Although some datasets like WebLogo-2M~\cite{Su2017WebLogo} and PL2K~\cite{II2019Scalable} are large, WebLogo-2M is very noisy for its unsupervised annotation while the recent proposed PL2K is not publicly available.

To address these problems, we introduce a new logo dataset, Logo-2K+ for logo classification. Compared with existing public available datasets, such as FlickrLogos-32, Logo-2K+ has three distinctive characteristics: (1) Large-scale. It consists of 167,140 images with a total number of 2,341 categories. (2) High-coverage. To our knowledge, Logo-2K+ consists of 2,341 categories and has the highest coverage of the logo categorical space.
(3) High-diversity. The logo images are collected from different websites. There are different types of logo images with different logo appearance, scale and background. Figure. \ref{fig1} shows some samples.

Furthermore,  we propose a Discriminative Region Navigation and Augmentation Network (DRNA-Net) to first localize informative logo regions, and then explores the potential of data augmentation for region-oriented data augmentation. The proposed DRNA-Net consists of four sub-networks: the navigator sub-network navigates the model to focus on informative regions. For each region, the teacher sub-network evaluates its probability belonging to ground-truth class to select the most informative ones. Region-oriented data augmentation subnetwork is mainly used to augment these selected regions via both region cropping and region dropping. The scrutinizer sub-network extracts and fuses features from augmented regions and the whole image to make logo classification. The four sub-networks benefit each other and can be trained in an end-to-end way.

In summary, this paper has three main contributions. First, we introduce a large-scale publicly available logo dataset Logo-2K+ with 2,341 logo classes and 167,140 images. To our knowledge, this is the largest publicly available high-quantity dataset for logo classification. Second, we propose a novel Discriminative Region Navigation and Augmentation Network (DRNA-Net) for scalable logo classification. It can automatically locate logo-relevant informative regions and conduct region-oriented data augmentation to extract more discriminative features from these augmented regions. Third, we conduct extensive evaluation on four datasets, including newly proposed Logo-2K+, and other three datasets with different scales, namely BelgaLogos~\cite{Neumann2001Integration}, FlickrLogos-32~\cite{Romberg2011Scalable} and WebLogo-2M~\cite{Su2017WebLogo}. The experimental results verified the effectiveness of the proposed method on all these datasets.

\section{Related Work}
In this section, we review related efforts towards (1) Logo-centric datasets, and (2) Logo classification.
\subsection{Logo-centric Datasets}
Several existing publicly available logo datasets have been used for classification, such as FlickrLogos-32 \cite{Romberg2011Scalable}, BelgaLogos \cite{Neumann2001Integration} and Logos in the wild \cite{Andras2017Open}. These datasets do not include a wide range of logo images and are lack of diversity and coverage in logo categories. Therefore, they are not sufficient to support complex statistical models, such as deep learning models for scalable logo classification. Although some larger datasets are proposed, such as LOGO-Net \cite{Hoi2015LOGO}, WebLogo-2M \cite{Su2017WebLogo} and PL2K \cite{II2019Scalable}. Unfortunately, most of these datasets are either very noisy or not publicly available. In contrast, our proposed Logo-2K+ is a large-scale high-coverage, high-quantity dataset with over two thousand logo categories and about 170 thousand logo images, where there are at least 50 images for one category. Table 1 summarizes the statistics of main logo datasets.
\begin{table}[!htbp]
\caption{Comparison between Logo-2K+ and existing logo datasets.}
\centering
\small
\renewcommand\arraystretch{1.05}
\setlength{\tabcolsep}{0.7mm}{
\begin{tabular}{c|c|c|c}
\hline
Dataset                                                                       & Logos       & Images             & Availability    \\ \hline
\begin{tabular}[c]{@{}c@{}}FlickLogos-27\\ \cite{Romberg2011Scalable}\end{tabular}                & 27               & 1,080              &${\surd}$           \\
\begin{tabular}[c]{@{}c@{}}FlickLogos-32\\ \cite{Romberg2011Scalable}\end{tabular}                & 32               & 8,240              & $\surd$            \\
\begin{tabular}[c]{@{}c@{}}BelgaLogos\\  \cite{Neumann2001Integration}\end{tabular}               & 37               & 10,000             & $\surd$            \\
\begin{tabular}[c]{@{}c@{}}FlickLogos-47\\ \cite{Romberg2011Scalable}\end{tabular}                & 47               & 8,240              & $\surd$            \\
\begin{tabular}[c]{@{}c@{}}LOGO-Net\\ \cite{Hoi2015LOGO}\end{tabular}                             & 160              & 73,414             & $\times$           \\
\begin{tabular}[c]{@{}c@{}}TopLogo-10   \\ \cite{Su2017Deep}\end{tabular}                         & 10               & 700                & $\surd$            \\
\begin{tabular}[c]{@{}c@{}}Logo-405\\ \cite{Hou2017Deep}\end{tabular}                             & 405              & 32,218             & $\times$           \\
\begin{tabular}[c]{@{}c@{}}Logos in the wild\\ \cite{Andras2017Open} \end{tabular}                & 871              & 11,054             & $\surd$            \\
\begin{tabular}[c]{@{}c@{}}WebLogo-2M\\ \cite{Su2017WebLogo}\end{tabular}                         & 194              & 1,861,177          & $\surd$            \\
\begin{tabular}[c]{@{}c@{}}PL2K\\ \cite{II2019Scalable}\end{tabular}                              & 2,000            & 295,814            & $\times$           \\
\textbf{Logo-2K+(Ours)}                                                        & \textbf{2,341} & \textbf{167,140}  & \textbf{$\surd$}\\ \hline
\end{tabular}}
\end{table}
\subsection{Logo Classification}
Logo classification has a long history in computer vision. Traditional logo classification methods resort to hand-crafted features, such as SIFT features and keypoint-based methods~\cite{Joly2009Logo,Romberg2013Bundle,Li2014Logo,Su2017WebLogo}. To improve the robustness of features, some methods adopt a symbolic representation of features \cite{Su2018Open} using global features for classification. Recently, deep learning methods \cite{Hoi2015LOGO,Iandola2015DeepLogo,Su2017Deep} have been proposed for logo classification. For example, Bianco \textit{et al.} \cite{Bianco2017Deep} propose a logo classification pipeline, which is composed of a logo region proposal followed by a Convolutional Neural Network (CNN) specifically trained for logo classification. In addition, there are also some works for logo detection~\cite{Su2017WebLogo,Oliveira2016Automatic,Li2018Graphic,Oliveira2016Automatic}, where the state-of-the-art deep neural detector models, such as Faster R-CNN are utilized.

Our work also focuses on logo classification. Different from previous work, our work mainly focuses on learning  discriminative logo-relevant regions in a self-supervised manner for logo classification. In addition, inspired by the spatial transformation for data augmentation~\cite{Hu2019See}, we introduce a region-oriented data augmentation strategy to further augment informative regions for more discriminative feature learning.

\section{Dataset: Logo-2K+}
\begin{figure}[!htbp]
\centering
\includegraphics[width=0.5\textwidth]{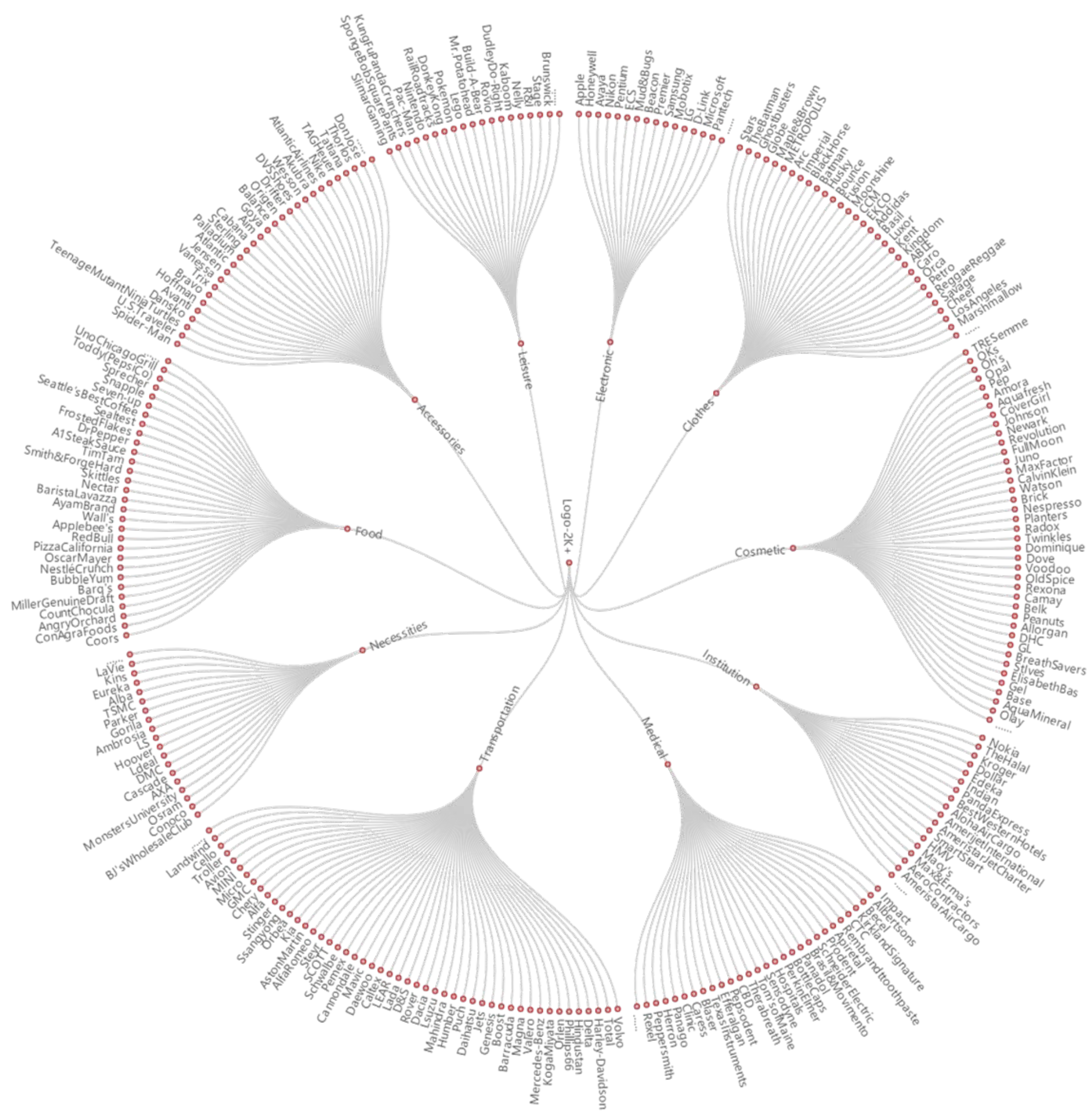} % Reduce the figure size so that it is slightly narrower than the column.
\caption{The boxable logo class hierarchy. Parent nodes represent more generic ten categories than their children (Best View under Magnification).}
\label{logo_class_hier}
\end{figure}
\begin{table}[!htbp]
\centering
\caption{Data statistics on Logo-2K+.}
\renewcommand\arraystretch{1.05}
\setlength{\tabcolsep}{2mm}{
% \begin{tabular}{c|c|c}
\begin{tabular}{p{2.8cm}<{\centering}|p{2cm}<{\centering}|p{2cm}<{\centering}}
\hline
Root Category    & \begin{tabular}[c]{@{}l@{}}Logos \end{tabular} & Images \\ \hline
Food             & 769 & 54,507 \\
Clothes          & 286 & 20,413 \\
Institution      & 238 & 17,103 \\
Accessories      & 210 & 14,569 \\
Transportation   & 203 & 14,719 \\
Electronic       & 191 & 13,972 \\
Necessities      & 182 & 13,205 \\
Cosmetic         & 115 & 7,929  \\
Leisure          & 99  & 7,338  \\
Medical          & 48  & 3,385  \\
\hline
Total & \textbf{2,341}& \textbf{167,140} \\ \hline
\end{tabular}}
\end{table}
As noted above, to date, there are no publicly available large-scale high-quantity logo datasets. In order to perform scalable logo classification, we construct a large-scale logo dataset, Logo-2K+, which covers a diverse range of logo classes from real-world logo images. The construction of Logo-2K+ is composed of two steps, from constructing the logo category list to collecting and cleaning logo images.

\textbf{Constructing the Logo Category List.} The first asset of a high-quality dataset should be a high-coverage of the categorical space.
We construct a logo list based on frequent appeared logo scenes and objects from the following 10 categories, namely \textbf{Food, Clothes, Institution, Accessories, Transportation, Electronic, Necessities, Cosmetic, Leisure, Medical.} For each root category, we further build a list of logo classes based on some logo websites, leading to the resulting 2,341 logo classes. Logo-2K+ has imbalanced class amounts in different categories. For example, there are 769 logo classes in the ``Food" category while only 48 classes in the ``Medical" category. Figure. \ref{logo_class_hier} shows the results of logo hierarchy visualization.

\textbf{Collecting and Cleaning Logo Images.} We crawl candidate images from the Google image search website. In order to obtain more relevant logo images, we expand search terms by adding keywords, such as ``logo" and ``brand". For example, ``Apple" means not only a famous brand of electronic products, but also a category of fruits. Therefore, we choose ``Apple logo" and ``Apple brand" as the search term to crawl logo images. We then manually check the quality of each class by removing duplicated images, images with terrible aspect ratio, too small logo ratio in the whole image. For logo class with less 50 images, we further enlarge the amount of logo images from the other websites, such as Baidu Images.

\begin{figure}[!htbp]
\centering
\includegraphics[width=0.5\textwidth]{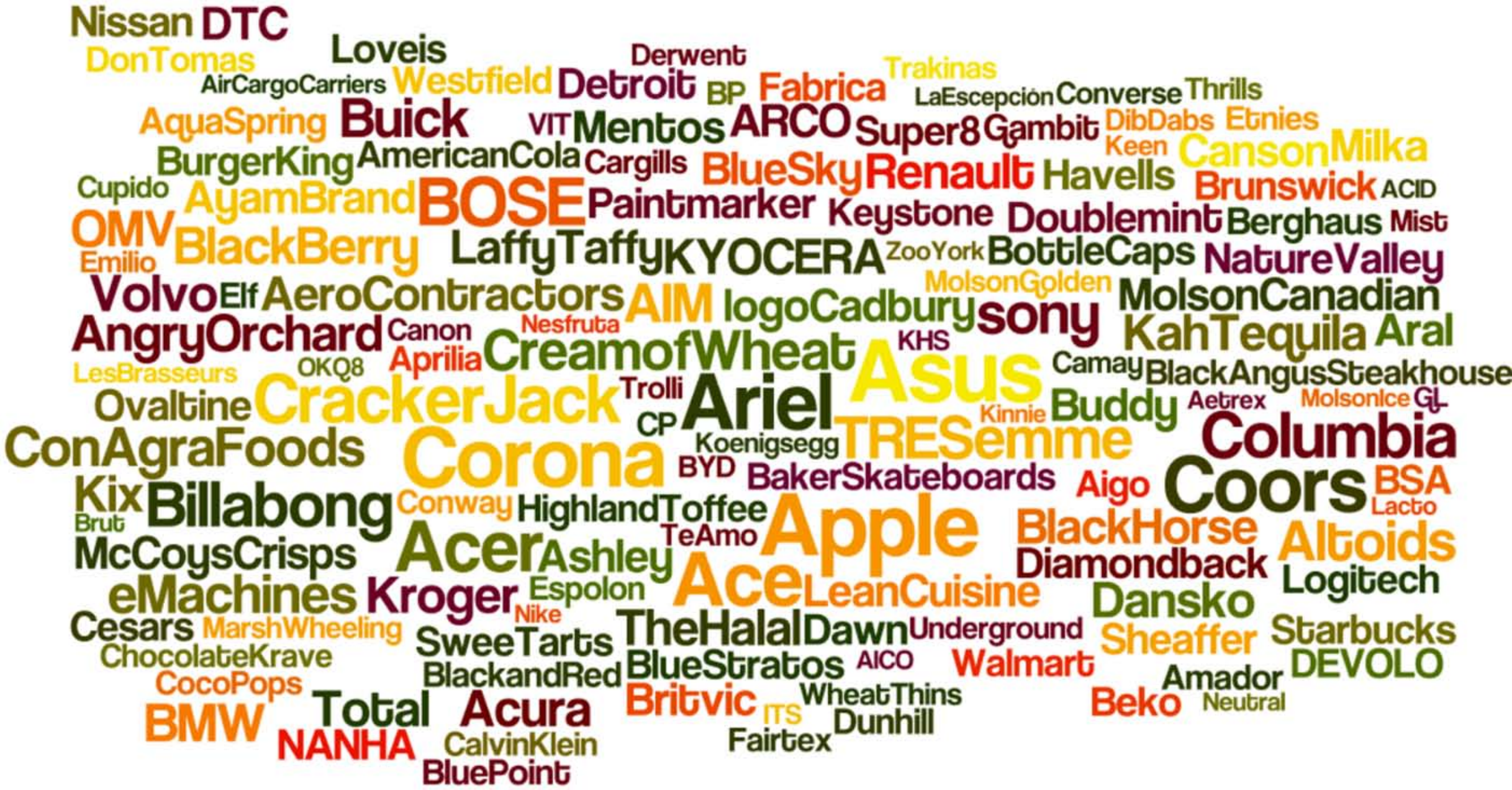} % Reduce the figure size so that it is slightly narrower than the column.
\caption{Visualization of logo classes based on the number of images in each class. Larger font size indicates more images in the corresponding class. Color is used randomly for visualization purpose only.}
\label{word_cloud}
\end{figure}

\textbf{Data Statistics.} Our resulting logo dataset contains 167,140 images with 10 root categories and 2,341 categories. There are at least 50 images for each logo class. The statistical comparison of 10 root categories from Logo-2K+ is shown in Table 2. Examples are shown in Figure \ref{fig1}. Figure \ref{word_cloud} shows the distribution of images across logo categories in Logo-2K+. We can see that the distribution of images for different logo categories is not uniform. Examples of logo categories with more images are Apple, Asus, Corona, to name a few. Examples with few images are Brub, Nike, KHS, to name a few.

\section{Approach}
\subsection{Overview}
\begin{figure*}[!htbp]
\centering
\includegraphics[width=1\textwidth]{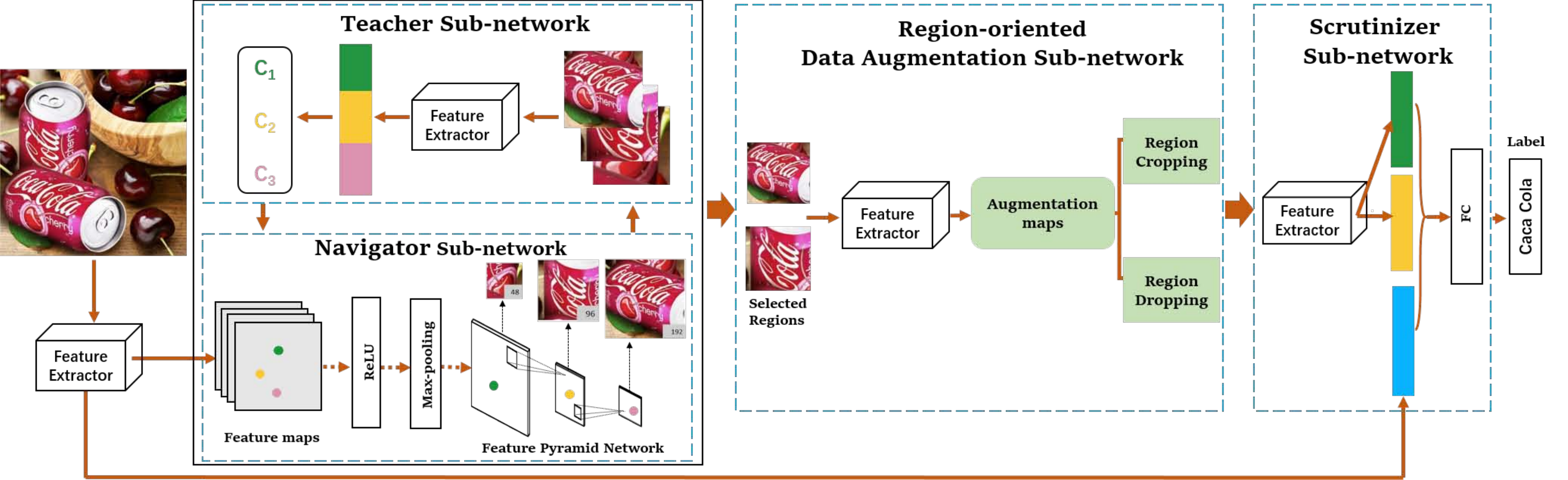} % Reduce the figure size so that it is slightly narrower than the column.
\caption{An overview of  proposed Discriminative Region Navigation and Augmentation Network (DRNA-Net). For the input image, these feature maps from the feature extractor are fed into the navigator sub-network to compute the informativeness of all regions. Then, these regions from the full image are also fed into the teacher sub-network to get the confidence. We optimize the navigator sub-network to make informativeness and confidence consistent. Next, we put selected regions with high informativeness into region-oriented data augmentation sub-network to augment these regions via both region cropping and region dropping. Finally, the scrutinizer network fuses features of those augmented regions and the full image to predict the label. FC denotes the full-connected layer.}
\label{DRNA-Net}
\end{figure*}
In this section, we will introduce the proposed Discriminative Region Navigation and Augmentation Network (DRNA-Net) for logo classification. Figure \ref{DRNA-Net} illustrates the architecture of DRNA-Net, which is mainly divided into four  parts, namely navigator sub-network, teacher sub-network, region-oriented data augmentation sub-network, and scrutinizer sub-network. Given an input image, the navigator sub-network first computes the informativeness of all regions generated from the pre-defined anchors. Then, the teacher sub-network evaluates each region's confidence that it belongs to the ground-truth class to enable the navigator sub-network to select the most informative logo-relevant regions. Next, the region-oriented data augmentation sub-network can augment the selected regions to localize more relevant logo regions. Finally, the scrutinizer sub-network fuses features from augmented regions and the whole image into the final feature representation for logo prediction.
\subsection{Navigator Sub-network}
Given an input image $X$, the navigator sub-network adopts convolutional layers, ReLU activation and max-pooling and a top-down feature pyramid network~\cite{Lin2016Feature} with lateral connections to generate regions $A$ with different scales and ratios. Each region $\begin{array}{l}R\in A\\\end{array}$. The non-maximum suppression on regions is used to reduce regional redundancy based on the informativeness, leading to top-M informative regions as $ R=\{R_1,R_2,\dots,R_M\}$ with corresponding informativeness $I=\{I_1,I_2,\dots,I_M\}$. These regions are then fed into the teacher sub-network to generate the most informative regions.

\subsection{Teacher Sub-network}
The teacher sub-network can calculate the confidence  $\left\{C(R_1),\;C(R_2),\dots,C(R_M)\right\}$ based on detected regions $ R=\{R_1,R_2,\dots,R_M\}$ from the navigator sub-network. The regions with a higher probability belonging to the ground-truth class should have higher confidence, that is
\begin{equation}\label{rank_condition}
\rm{if}\;C(R_1)>C(R_2),\; \rm{then}\;I(R_1)>I(R_2);R_1,\;R_2\in A
\end{equation}
We optimize the navigator sub-network and teacher sub-network to make $\left\{C(R_1),\;C(R_2),\dots,C(R_M)\right\}$ and $\left\{I(R_1),\;I(R_2),\dots,I(R_M)\right\}$ keep the same order using Eq.~\ref{rank_condition} via the following pair-wise ranking loss between predicted confidence and the ground-truth class:
\begin{equation}
Loss_I(I,C)={\textstyle\sum_{(i,j):C_i<C_j}}f(I_j-I_i)
\end{equation}
where $i,j$ are the region index and $f(x)=\rm{max}\{1-x,0\}$. $Loss_I(I,C)$ encourages if $C_i<C_j$, then $I_i<I_j$. The loss function enforces consistency between informativeness of region and probability being ground-truth class.

Then, the teacher sub-network $Loss_c$ is defined to minimize the difference between losses of all regions $\log C(R_i)$ and the full image $\log C(X)$ as follows:
\begin{equation}
Loss_c=-{\textstyle\sum_{i=1}^M}\log C(R_i)-\log C(X)
\end{equation}

\subsection{Region-oriented Data Augmentation Sub-network}
We obtain the most informative logo-relevant regions $\left\{C(R_1),\;C(R_2),\dots,C(R_K)\right\}$, where $K$ is the number of selected regions through both the navigator sub-network and teacher sub-network.  Region-oriented data augmentation sub-network is then introduced to augment these selected regions to obtain more relevant regions via both region cropping and region dropping~\cite{Hu2019See}. We first normalize $R_K$ to generate augmentation maps $R_k^\ast$ as follows:
\begin{equation}
R_k^\ast=\frac{R_k-min\left(R_k\right)}{max\left(R_k\right)-min\left(R_k\right)}
\end{equation}

(1) Region cropping: with augmentation map $R_k^\ast$, we can zoom into the region and extract more informative local feature. We set a threshold $\theta_c\in\left[0,1\right]$ to determine whether the element $R_k^\ast\left(i,j\right)$ belongs to 0 or 1, the formula as follows,
\begin{equation}
C_k^{}\left(i,j\right)=\left\{\begin{array}{l}1,\;\;\;\;\;\;if\;R_k^\ast\left(i,j\right)>\theta_c\\0,\;\;\;\;\;\;\;\rm{otherwise}\end{array}\right.
\end{equation}
where $C_k^{}$ is the 0-1 crop mask, and a bounding box on this region can be found to cover the selected positive location of $C_k^{}$.

(2) Region dropping: region dropping is used to encourage region maps to represent multiple discriminative logo' parts, and is formulated as follows:
\begin{equation}
D_k^{}\left(i,j\right)=\left\{\begin{array}{l}1,\;\;\;\;\;\;if\;R_k^\ast\left(i,j\right)>\theta_d\\0,\;\;\;\;\;\;\;\rm{otherwise}\end{array}\right.
\end{equation}

The $k_{th}$ selected region will be dropped by masking region with $D_k$.

We denote $E$ as the augmentation function and adopt cross entropy $Loss_a$ as follows,
\begin{equation}
Loss_a=-{\textstyle\sum_{i=1}^K}\log E(R_k^\ast)-\log E(X)
\end{equation}
where  $E$  maps the augmentation selected region to its probability being ground-truth class label.

\subsection{Scrutinizer Sub-network}
The scrutinizer sub-network obtains $K$ augmented informative regions, and uses a feature extractor to extract and fuse the features of the full image $X$ and regions $\{R_1^\ast,\;R_2^\ast,\;...,\;R_K^\ast\}$. We concatenate  $K$ augmentation features vector and input image $X$ feature vector, and feed them into the full convolution layer. The scrutinizer sub-network finally gives the prediction results $P=F(X,\;R_1^\ast,\;R_2^\ast,\;...,\;R_K^\ast)$, where $F$ is a transformation, we define the classification cross entropy loss $ Loss_s$ is used.

\subsection{Joint Training Loss}
Finally, we use stochastic gradient descent method to optimize the total loss as follows:
\begin{equation}
L_{total}=Loss_I+\alpha\cdot Loss_s+\beta\cdot Loss_c+\gamma\cdot Loss_a
\label{loss_Total}
\end{equation}
where $\alpha$, $\beta$ and $\gamma$ are hyper parameters.

DRNA-Net is inspired by work \cite{Yang2018Learning}, but introduced  region-oriented data augmentation sub-network for selected regions to further enhance final feature representation. Therefore, DRNA-Net achieves two-level coarser-to-finer localization to efficiently discover more discriminative regions better for logo classification.
\section{Experiments}

\subsection{Experimental Setup}
For our method, we adopt ResNet-50 and ResNet-152 pre-trained on ILSVRC2012 as the feature extractor. The thresholds of region cropping and dropping $\theta_c$ and $\theta_d$ are both set to 0.5. We empirically set $M=4$ in the navigator sub-network and $K=2$ in the teacher sub-network. In Eq.~\ref{loss_Total}, hyper-parameters weights $\alpha=\beta=\gamma=1$ without prior. It is optimized using the stochastic gradient descent with a momentum of 0.9, a batch size of 8 and weight decay of 0.0001. We adopt the Pytorch framework to train the network and implement our algorithm on a NVIDIA Tesla V100 GPU (32GB). Both Top-1 accuracy and Top-5 accuracy are adopted as evaluation metrics.

\subsection{Experiments on Logo-2K+}
\subsubsection{Implementation Detials.}
70\%, 30\% of images are randomly selected for training and testing in each logo category. All the models are trained for 100 epochs with an initial learning rate of 0.001 and decreased after 20 epochs to 0.0001.

For baselines,  we evaluate different networks on Logo-2K+, such as AlexNet \cite{Alex2012ImageNet}, GoogLeNet \cite{Christian2014Going}, VGGNet \cite{Karen2014Very} and ResNet \cite{He2016Deep} for logo classification. We also list some efficient training and optimization method named Label Smoothing Regularization (LS)~\cite{He-BOT-CoRR2018}.

\begin{table}[t]
\caption{Comparison of our model and baselines on Logo-2K+ (\%).}
\centering
\renewcommand\arraystretch{1.05}
\setlength{\tabcolsep}{1.5mm}{
\begin{tabular}{c|c|c}
\hline
Method                                                                                    & Top-1 Acc.     & Top-5 Acc.     \\ \hline
AlexNet                                                                                   & 48.80          & 78.45          \\
GoogLeNet                                                                                 & 62.36          & 88.33          \\
VGGNet-16                                                                                 & 62.83          & 89.01          \\
ResNet-50                                                                                 & 66.34          & 91.01          \\
ResNet-152                                                                                & 67.65          & 91.52          \\
\begin{tabular}[c]{@{}c@{}}VGGNet-16+Efficient+LS\\ \cite{He-BOT-CoRR2018}\end{tabular}   & 65.45          & 90.12          \\
\begin{tabular}[c]{@{}c@{}}ResNet-50+Efficient+LS\\ \cite{He-BOT-CoRR2018}\end{tabular}   & 66.94          & 91.30          \\
\begin{tabular}[c]{@{}c@{}}ResNet-152+Efficient+LS\\\cite{He-BOT-CoRR2018}\end{tabular}   & 67.99          & 91.68          \\
\begin{tabular}[c]{@{}c@{}}NTS-Net(ResNet-50)\\\cite{Yang2018Learning}\end{tabular}       & 69.41          & 91.95          \\
DRNA-Net(ResNet-50)                                                                       & 71.12          & 92.33          \\
DRNA-Net(ResNet-152)                                                                      & \textbf{72.09} & \textbf{93.45} \\ \hline
\end{tabular}}
\label{result_com_Logo2K}
\end{table}

The evaluation results on Logo-2K+ dataset are shown in Table \ref{result_com_Logo2K}. We can see that (1) The best single network is ResNet-152, with a Top-1 accuracy of 67.65\% and a Top-5 accuracy of 91.52\%, and achieves a 67.99\% Top-1 accuracy using some tricks. (2) Our method produces the best $72.09\%$ in Top-1 accuracy and $93.45\%$ in Top-5 accuracy, surpassing the NTS-Net about $1\%$. This indicts that the  effectiveness of region-oriented data augmentation strategy.
\begin{figure}[!htbp]
\centering
\includegraphics[width=0.47\textwidth]{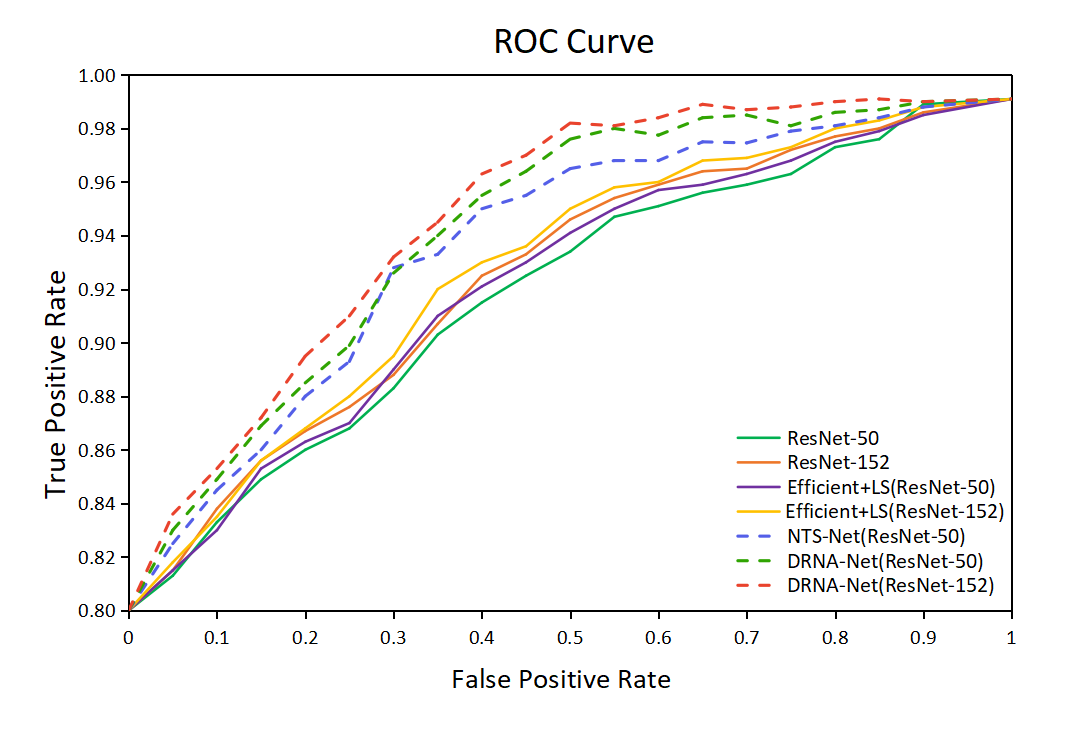}
\caption{ROC curves of different logo classification models on Logo-2K+.}
\label{ROC}
\end{figure}

Considering image samples from different logo classes are unbalanced, to further comprehensively evaluate the performance of DRNA-Net, we further draw the ROC curves of all the methods in Figure~\ref{ROC}, where the dotted green and red curve represent the performance of our method and the purple curve refers to the NTS-Net method. We can see that the true positive rate remains high on the DRNA-Net compared to NTS-Net for test logo classes,
\begin{figure}[!htbp]
\centering
\includegraphics[width=0.48\textwidth]{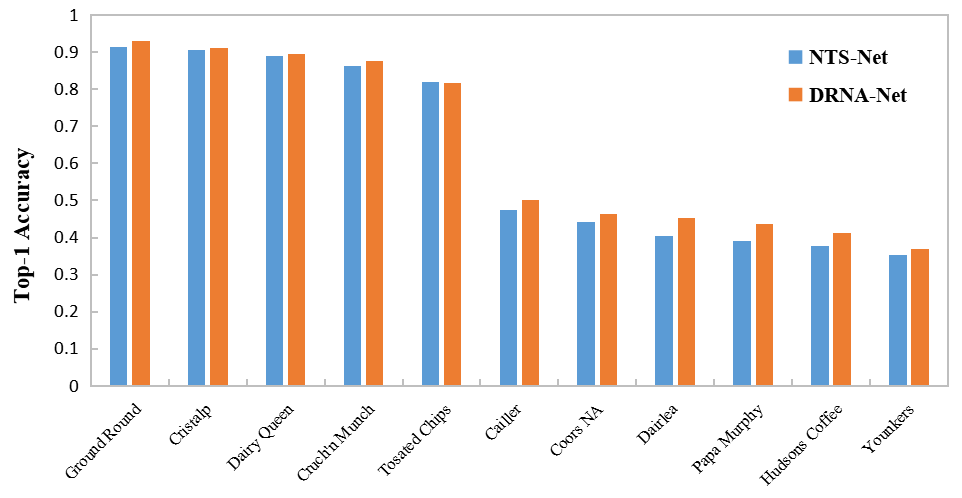}
\caption{Selected classification Top-1 Accuracy: The 5 best and 5 worst performing classes are shown.}
\label{Top5_example}
\end{figure}
and DRNA-Net obtains the best performance in terms of overall quality for logo classification.

In addition, we select 10 classes in the test phase to further evaluate our method and NTS-Net on the ResNet-50 base CNN. Particularly, we listed the Top-1 Accuracy of both 5 best and 5 worst performing classes in Figure \ref{Top5_example}. As shown in Figure \ref{Top5_example}, we find that for 9 classes among 10 logo classes, there is the performance improvement for our method and the performance from 1 additional classes is degraded compared to NTS-Net. The improvements for single classes are visibly more pronounced than degradations, indicating the effectiveness of introduced region-oriented data augmentation sub-network. In addition, for the categories with bad classification results, we observe that most of these logo classes contain fewer samples, or the logo images contain  the smaller logo region  or complex background.
\subsubsection{Visualization of Localized Regions.}
\begin{figure}[!htbp]
\centering
\includegraphics[width=0.475\textwidth]{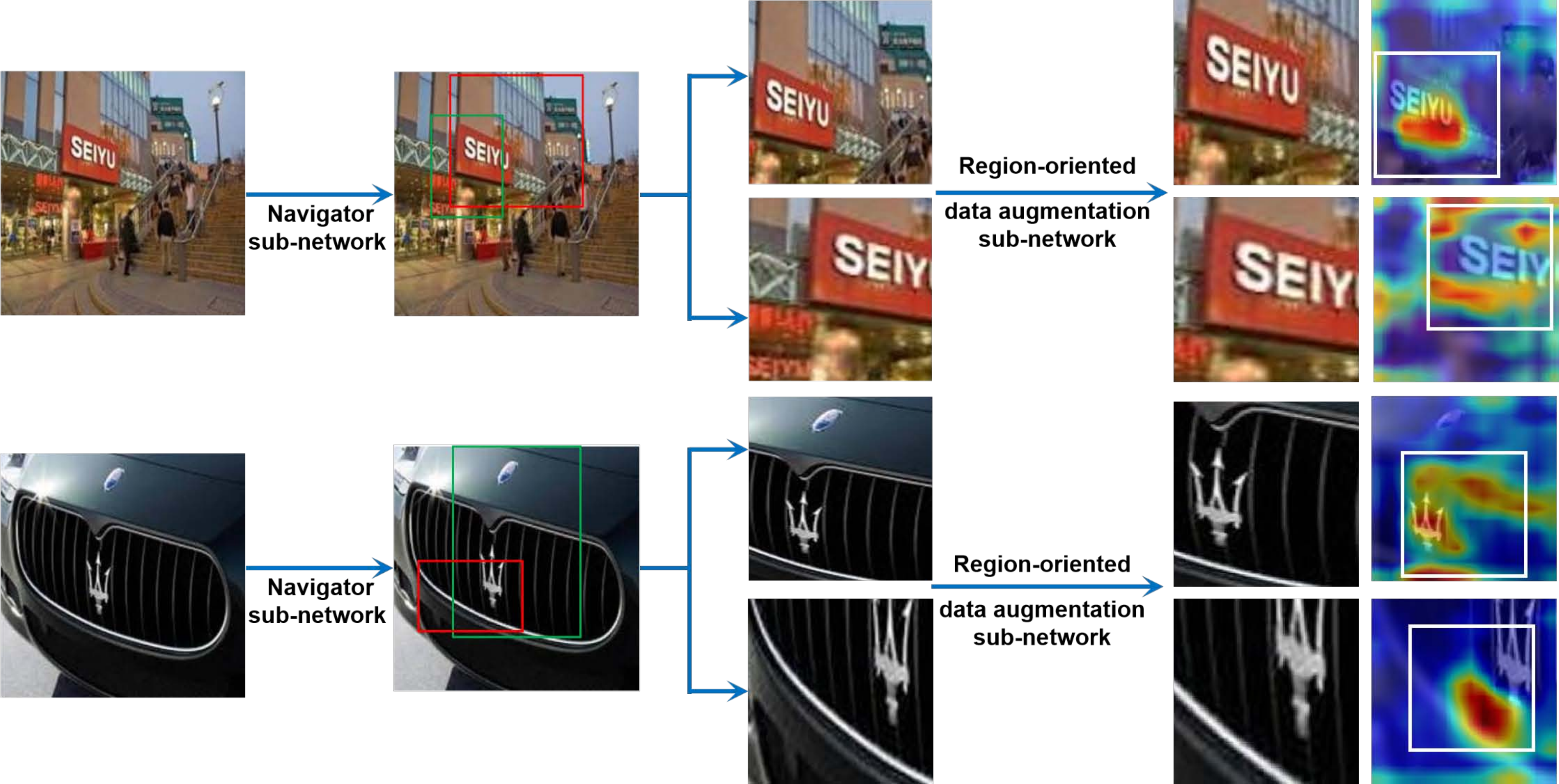}
\caption{Examples of the visualization of discovered regions by DRNA-Net.}
\label{visual}
\end{figure}

To analyze the localization effect of our method, we visualize the regions from some logo samples by navigator sub-network and region-oriented data augmentation sub-network in Figure \ref{visual}. First, we adopt the navigator sub-network to coarser localization regions guided by teacher sub-network. Then, the regions are feed into the region oriented data augmentation sub-network to achieve the finer localization. We can see that after two-level coarser-to-finer localization, our method can  find more relevant logo regions to support logo classification.
\subsection{Experiments on Other Datasets }
Besides Logo-2K+, we also conduct the evaluation on another publicly available benchmark datasets, BelgaLoges, FlickrLogo-32 and WebLogo-2M to further verify the effectiveness of our method. \textbf{BelgaLoges} contains 37 logo class and a total of 10,000 images. It is a small-scale logo image dataset for product brand recognition. \textbf{FlickrLogos-32} contains 32 different logo brands, 8,240 logo images. \textbf{WebLogo-2M} contains 1,867,177 logo images in 194 different logo classes. This dataset is annotated via an unsupervised process, and thus is very noisy. For all three datasets, 70\%, 30\% of images are randomly selected for training and testing for each logo category.
\begin{table}[t]
\caption{Performance comparison of methods on BelgaLoges(\%).}
\centering
\renewcommand\arraystretch{1.1}
\setlength{\tabcolsep}{1.5mm}{
\begin{tabular}{c|c|c}
\hline
Method                                                                                         & Top-1 Acc.     & Top-5 Acc.      \\ \hline
\begin{tabular}[c]{@{}c@{}}RCNN(CaffeNet)\\ \cite{Girshick2014Rich}\end{tabular}               & 91.80          & -               \\
\begin{tabular}[c]{@{}c@{}}FRCN(VGGNet-16)\\ \cite{Girshick2015Fast}\end{tabular}              & 87.30          & -               \\
\begin{tabular}[c]{@{}c@{}}SPPnet(ZF)\\ \cite{He2014Spatial}\end{tabular}                      & 87.70          & -               \\
\begin{tabular}[c]{@{}c@{}}NTS-Net(ResNet-50)\\ ~\cite{Yang2018Learning}\end{tabular}          & 93.33          & 96.15           \\
DRNA-Net(VGGNet-16)                                                                            & 92.41          & 95.96           \\
DRNA-Net(ResNet-50)                                                                            & 94.44          & 97.11           \\
DRNA-Net(ResNet-152)                                                                           & \textbf{95.82} & \textbf{98.40}  \\ \hline
\end{tabular}}
\label{BelgaLoges_results}
\end{table}
\subsubsection{Experiments on BelgaLoges}
We list experimental results of baselines and our proposed method in Table \ref{BelgaLoges_results}. These models are trained for 80 epochs with an initial learning rate of 0.001 and divided by 10 after 20 epochs. As we can see from Table \ref{BelgaLoges_results}, our method achieves the best performance: the $94.44\%$ Top-1 accuracy with ResNet-50 and the $95.82\%$ Top-1 accuracy with ResNet-152. In addition, we also compare the FRCN (VGGNet-16) with our method with VGGNet-16, and find that there is about $5\%$ improvement in Top-1 accuracy. These experimental results can further  demonstrate the effectiveness of our proposed method.
\subsubsection{Experiments on FlickrLogos-32}
All models are trained for 80 epochs with an initial learning rate of 0.001 and divided by 10 after 20 epochs, and the batch size is 32. The classification accuracy of different methods is summarized in Table \ref{FlickrLogos_results}. We can see that our method achieves the best performance in both Top-1 accuracy and Top-5 accuracy. Compared with NTS-Net, the proposed strategy obtains 1.2\% improvement, which demonstrates that introducing region-oriented data augmentation method can help improve the performance.
\begin{table}[!htbp]
\caption{Comparison of our model and state-of-the-art methods on FlickrLogos-32 (\%).}
\centering
\renewcommand\arraystretch{1.1}
\setlength{\tabcolsep}{1.5mm}{
\begin{tabular}{c|c|c}
\hline
Method                                                                                                & Top-1 Acc.     & Top-5 Acc.  \\ \hline
\begin{tabular}[c]{@{}c@{}}FRCN + AlexNet\\ \cite{Iandola2015DeepLogo}\end{tabular}                   & 75.00          & -              \\
\cite{Eggert2015On}                                                                                   & 84.60          & -              \\
\cite{Bianco2015Logo}                                                                                 & 88.40          & -              \\
\cite{Bianco2017Deep}                                                                                 & 91.70          & -              \\
\begin{tabular}[c]{@{}c@{}}SIFT\\ \cite{Romberg2013Bundle}\end{tabular}                               & 94.10          & -              \\
\begin{tabular}[c]{@{}c@{}}NTS-Net(ResNet-50)\\ ~\cite{Yang2018Learning}\end{tabular}                 & 94.14          & 96.29          \\
DRNA-Net(ResNet-50)                                                                                   & 95.33          & 97.17          \\
DRNA-Net(ResNet-152)                                                                                  & \textbf{96.63} & \textbf{98.80} \\ \hline
\end{tabular}}
\label{FlickrLogos_results}
\end{table}
\subsubsection{Experiments on WebLogo-2M}
In order to better prove the effectiveness of our method, we also carry out experiments on the WebLogo-2M dataset with large-scale but noisy images. The some experimental setup are the same as the settings of Logo-2K+. The experimental results of baselines and our method is summarized in Table \ref{WebLogo_results}. We can see that our method achieves the best performance with ResNet-50 compared with other baselines, such as NTS-Net. We can also see that although there are only 194 logo classes WebLogo-2M with many images for each logo class, the performance on WebLogo-2M is lower. The probable reason is that there are many noisy images in the WebLogo-2M.
\begin{table}[!htbp]
\caption{Comparison of our model and baselines on WebLogo-2M  (\%).}
\centering
\renewcommand\arraystretch{1.1}
\setlength{\tabcolsep}{1.5mm}{
\begin{tabular}{c|c|c}
\hline
Method                                                                                & Top-1 Acc.     & Top-5 Acc.     \\ \hline
AlexNet                                                                               & 50.99          & 70.62          \\
GoogLeNet                                                                             & 62.14          & 80.21          \\
VGGNet-16                                                                             & 62.88          & 83.23          \\
ResNet-50                                                                             & 62.93          & 83.32          \\
\begin{tabular}[c]{@{}c@{}}NTS-Net(ResNet-50)\\ \cite{Yang2018Learning}\end{tabular}  & 63.67          & 84.31              \\
DRNA-Net(ResNet-50)                                                                   & \textbf{64.82}          & \textbf{86.12}          \\
\hline
\end{tabular}}
\label{WebLogo_results}
\end{table}
\section{Conclusion}
In this paper, we have introduced a novel large-scale logo benchmark dataset Logo-2K+ with over 2000 logo classes and about 170,000 logo images. This dataset should further the state of the art in scalable logo image recognition. As one strong baseline, we have also presented a method to discover and augment discriminative regions to enhance final feature representation, and have shown its effectiveness on Logo-2K+ and other three existing logo benchmarks. Future work includes conducting the bounding-box annotation of logos in logo images or synthesizing logo-annotated images without expensive manual labeling~\cite{Su2017Deep} to support large-scale logo detection task based on Logo-2K+.

\bibliography{bibfile111}
\bibliographystyle{aaai}
\end{document}